\documentclass[3p,times,procedia]{elsarticle}
\usepackage{ecrc}
\usepackage{subcaption}
\usepackage{booktabs}
\usepackage{amsmath}
\usepackage{amsthm}
\usepackage[bookmarks=false]{hyperref}
    \hypersetup{colorlinks,
      linkcolor=blue,
      citecolor=blue,
      urlcolor=blue}

\volume{00}

\firstpage{1}

\journalname{Procedia Computer Science}

\runauth{Author name}

\jid{procs}

\usepackage{amssymb}
\usepackage{multirow}
\usepackage{wrapfig}

\newtheorem{theorem}{Theorem}

\usepackage[figuresright]{rotating}

\usepackage[figuresright]{rotating}

\begin{document}
\begin{frontmatter}



\dochead{The 11th International Conference on Computer Science and Computational Intelligence 2026}

\title{Quantifying How Training Gradient Sparsity Affect Spiking Neural Network Accuracy And Robustness}


\author[a]{Nhan Trong Luu\corref{cor1}} 
\author[b]{Duong Trung Luu}

\address[a]{College of ICT, Can Tho University, Can Tho, Vietnam}
\address[b]{Center of Digital Transformation and Communication, Can Tho University, Can Tho, Vietnam}

\begin{abstract}
Spiking Neural Networks (SNNs) have recently received increasing attention in both computational neuroscience and artificial intelligence owing to their potential for energy-efficient computation and reduced memory requirements. Despite these advantages, improving adversarial robustness in SNNs (particularly for vision-based applications) remains an emerging and relatively underexplored research problem. Recent work has suggested that encouraging sparse gradients can act as a regularization mechanism to improve resistance against adversarial perturbations. In this study, we report an unexpected observation: under certain architectural configurations, SNNs inherently exhibit sparse gradients and can attain state-of-the-art adversarial defense performance without requiring any explicit regularization strategy. Further investigation reveals an inherent trade-off between robustness and generalization. Specifically, increased gradient sparsity enhances resistance to adversarial attacks but may reduce the model's generalization capability, whereas denser gradients tend to improve generalization while simultaneously increasing susceptibility to adversarial perturbations. These findings provide new perspectives on the role of gradient sparsity in the training dynamics of SNNs.
\end{abstract}

\begin{keyword}
Spiking neural network; gradient sparsity; adversarial robustness; image classification




\end{keyword}
\cortext[cor1]{Corresponding author.}
\end{frontmatter}

\email{luutn@ctu.edu.vn}

\section{Introduction}

Recent progress in artificial intelligence (AI) has been largely fueled by the rapid development of deep learning techniques \cite{luu2026robust}. Despite their success, conventional artificial neural networks (ANNs) capture only limited aspects of the complex temporal dynamics present in biological neural systems \cite{luu2026direct}. Spiking neural networks (SNNs) aim to bridge this gap by modeling neural computation through discrete spike events, thereby enabling temporally driven processing that more closely resembles biological information transmission.

In SNNs, neurons communicate through asynchronous spike signals, reflecting the event-driven nature of biological neural activity \cite{11261679}. This communication paradigm enhances biological realism while also offering practical advantages such as improved energy efficiency and compatibility with neuromorphic hardware platforms \cite{luu2024}. Nevertheless, training SNNs remains challenging because spike generation is inherently non-differentiable, preventing the direct use of standard backpropagation algorithms \cite{luu2026survey}. To address this issue, various learning approaches have been proposed, including biologically inspired spike-timing-dependent plasticity (STDP) \cite{luu2026parameter} and surrogate gradient methods that approximate gradients through non-differentiable spike functions \cite{luu2026survey}.

Robustness is another critical consideration in modern AI systems, particularly in safety-critical scenarios such as autonomous driving, where adversarial perturbations can lead to highly confident yet incorrect predictions \cite{szegedy2013intriguing, goodfellow2014explaining}. Although numerous defense strategies have been proposed \cite{ross2017right, madry2017towards}, neural networks across different architectures remain susceptible to adversarial manipulation.

Compared with ANNs, SNNs process temporally encoded spike patterns, which introduces unique characteristics in adversarial settings \cite{sharmin2020inherent}. Previous studies have suggested that certain rate-based SNN conversion approaches exhibit inherent robustness \cite{sharmin2020inherent, kundu2021hire}, while other work has demonstrated improved adversarial resistance through adversarial regularization when direct spike coding is employed \cite{ding2022snn, liu2024enhancing}. However, adversarial properties in SNNs trained with temporally coded inputs remain relatively underexplored.

Another important aspect of robustness is the well-known accuracy–robustness trade-off \cite{tsipras2018robustness}. While this phenomenon has been extensively investigated in ANNs, its underlying mechanisms in SNNs are not yet well understood. In this work, we examine how \emph{gradient sparsity} influences the robustness–generalization trade-off in temporally coded SNNs subjected to adversarial perturbations. Our main contributions are summarized as follows:
\begin{itemize}
    \item \textbf{Natural robustness of temporally trained SNNs:} Temporally coded SNNs optimized with surrogate gradients exhibit inherent resistance to adversarial perturbations, achieving strong robustness without requiring explicit defense strategies or auxiliary mechanisms \cite{ding2022snn, kundu2021hire, sharmin2020inherent}.
    
    \item \textbf{Gradient sparsity governs the trade-off:} Through both empirical observations and theoretical analysis, we show that shared computational pathways between input gradients and weight gradients induce gradient sparsity. This sparsity reduces model expressivity and weakens adversarial gradient signals, leading to lower clean accuracy but stronger adversarial robustness. Conversely, increasing gradient density improves generalization performance while simultaneously increasing vulnerability to adversarial attacks.
\end{itemize}

To the best of our knowledge, this study provides the first systematic analysis identifying gradient sparsity as a key factor governing the robustness–accuracy trade-off in SNNs.

\section{Adversarial attacks}

We briefly introduce the notation and adversarial attack methods used as evaluation baselines. Consider a differentiable classifier $f_\theta(X)$ parameterized by $\theta$, which produces predictions $\hat{y} \in \mathbb{R}^{N \times K}$ for input samples $X \in \mathbb{R}^{N \times D}$ across $K$ classes. Model training minimizes the cross-entropy loss between predicted outputs and one-hot encoded labels $y$ where $H(\cdot,\cdot)$ denotes the cross-entropy loss:
\begin{equation}
\theta^\ast = \arg\min_\theta \sum_{n=1}^{N}\sum_{k=1}^{K}
- y_{nk}\log f_\theta(X_n)_k
= \arg\min_\theta H(y,\hat{y}).
\label{eq:cel}
\end{equation}

\subsection{Fast Gradient Sign Method (FGSM)}

The Fast Gradient Sign Method (FGSM) \cite{goodfellow2014explaining} constructs adversarial samples by perturbing the input in the direction of the sign of the gradient with respect to the input. Specifically, adversarial examples are generated as $X_{\text{FGSM}} =
X + \epsilon \cdot
\text{sign}\big(\nabla_x H(y,\hat{y})\big)$, where $\epsilon$ controls the magnitude of the perturbation. Iterative extensions of this method apply multiple smaller gradient-based updates to strengthen the attack \cite{kurakin2018adversarial}.

\subsection{Projected Gradient Descent (PGD)}

Projected Gradient Descent (PGD) \cite{madry2017towards} generates adversarial examples through iterative gradient-based perturbations while enforcing a constraint on the perturbation magnitude. At each iteration, the input is updated as $X^{t+1} =
\Pi_{\mathcal{B}_\epsilon(X)}
\left(
X^t + \alpha \cdot
\text{sign}\big(\nabla_x H(y,\hat{y})\big)
\right)$, where $\alpha$ denotes the step size and $\Pi_{\mathcal{B}_\epsilon(X)}$ represents projection onto the $\ell_\infty$ ball of radius $\epsilon$ centered at the original input $X$. Repeating this process over multiple iterations produces stronger first-order adversarial examples while ensuring that perturbations remain bounded.

\section{Evaluation and analysis}

\subsection{Overall settings}

For the undefended benchmark, we adopted SEW-ResNet \cite{fang2021deep} and Spiking-ResNet \cite{hu2021spiking} architectures. Model implementation and experimental workflows were developed using PyTorch \cite{paszke2019pytorch} together with the SpikingJelly framework \cite{doi:10.1126/sciadv.adi1480}. Network optimization was performed using BPTT \cite{ding2022snn, fang2021incorporating} with surrogate gradients \cite{fang2021deep} applied to all IF neurons. Specifically, we employed the arctan surrogate function \cite{fang2021deep, eshraghian2023training}. The derivative $\partial \Theta/\partial x$ for the Heaviside function $\Theta(x, V_{th})$ with threshold voltage $V_{th}$ and input $x$ during backpropagation is defined as
\begin{equation}
\Theta(x, V_{th}) =
\begin{cases}
1 & \text{if } x - V_{th} \geq 0 \\
0 & \text{otherwise}
\end{cases}, \quad
\frac{\partial \Theta}{\partial x} =
\frac{\alpha}{2\left( 1 + \left( \frac{\pi}{2} \alpha x \right)^2 \right)}
\quad \text{where } \alpha=2.
\label{heviside}
\end{equation}

For spike encoding, we utilized the commonly adopted \textit{phase coding} scheme \cite{kim2018deep} with a time horizon of $T = 10$. The spike weight $\omega(t)$ for timestep $t \in T$ is defined as $\omega(t) = 2^{-(1 + \operatorname{mod}(t - 1, K))}$. Consistent with established practices in ANN and SNN optimization \cite{luu2023blind, luu2024, 11261679, luu2024universal}, model training employed the cross-entropy loss (see Equation~\ref{eq:cel}) together with the Adam optimizer. The learning rate was set to $lr = 1 \times 10^{-3}$ with optimizer parameters $\beta = (0.9, 0.999)$ across all experiments. Training was performed for 100 epochs using a batch size of 16 images on an NVIDIA RTX 3090 GPU. To ensure reproducibility and mitigate issues related to stochastic gradients \cite{athalye2018obfuscated}, deterministic random seeds were applied.

\subsection{Natural adversarial robustness of SNN over temporal-coded backpropagation}
\label{benchmark}

We evaluated the selected models on CIFAR10, CIFAR100 \cite{alex2009learning}, and CIFAR10-DVS \cite{li2017cifar10}, comparing them against several SOTA approaches designed to improve adversarial robustness in SNNs. In particular, our undefended SNN models were compared with input discretization methods \cite{sharmin2020inherent}, Hire-SNN \cite{kundu2021hire}, and SNN-RAT \cite{ding2022snn}, all of which have demonstrated strong performance under adversarial conditions. The evaluation focused on standard adversarial attacks, namely FGSM and PGD, with all models tested under identical attack configurations. Specifically, we set $\epsilon = 8/255$ for both FGSM and PGD, while PGD additionally used a step size $\alpha = 0.01$ and $t = 7$ iterations, following the configuration adopted in previous work \cite{ding2022snn}.

\begin{table*}[t!]
    \caption{Comparison of validation accuracy of undefended SNN variants with various SNN adversarial defense methods. Best performance per attack are denoted in bold. Notably, undefended SEW-ResNet achieve SOTA performance against PGD attacks and reasonable performance against FGSM attacks.}
    \begin{subtable}{0.49\textwidth}
    \centering
    \caption{Various SNN adversarial defense methods vs clean models on CIFAR-100.}
    \resizebox{\textwidth}{!}{
    \begin{tabular}{l*{3}{c}} 
    \toprule
     \multirow{2}{*}{SNN variants} & \multicolumn{3}{c}{Validation accuracy per attack (\%)}\\
    \cmidrule{2-4}
     & FGSM & PGD & Clean\\
    \midrule
    Input discretization \cite{sharmin2020inherent}$^{\dagger}$ & 15.50 & 6.30 & 64.40\\
    \midrule
    Hire-SNN \cite{kundu2021hire}$^{\dagger}$ & 22.00 & 7.50 & 65.10\\
    \midrule
    Vanilla VGG-11 \cite{ding2022snn}$^{\dagger}$ & 5.30 & 0.02 & \textbf{73.33}\\
    \midrule
    SNN-RAT VGG-11 \cite{ding2022snn}$^{\dagger}$ & 25.86 & 10.38 & 70.89\\
    \midrule
    SNN-RAT WRN-16 \cite{ding2022snn}$^{\dagger}$ & \textbf{28.08} & 11.31 & 69.32\\
    \midrule
    SEW-ResNet18 \cite{fang2021deep} & 20.81~$\pm$~0.523 & 13.47~$\pm$~0.246 & 42.19~$\pm$~0.373\\
    \midrule
    SEW-ResNet34 \cite{fang2021deep} & 19.91~$\pm$~0.427 & 13.25~$\pm$~0.382 & 41.81~$\pm$~0.292\\
    \midrule
    SEW-ResNet50 \cite{fang2021deep} & 21.27~$\pm$~0.272 & \textbf{16.03~$\pm$~0.292} & 38.91~$\pm$~0.387\\
    \midrule
    Spiking-ResNet18 \cite{hu2021spiking} & 13.36~$\pm$~0.285 & 11.15~$\pm$~0.254 & 23.23~$\pm$~0.258\\
    \midrule
    Spiking-ResNet34 \cite{hu2021spiking} & 2.434~$\pm$~0.127 & 1.643~$\pm$~0.178 & 21.20~$\pm$~0.197\\
    \midrule
    Spiking-ResNet50 \cite{hu2021spiking} & 3.326~$\pm$~0.165 & 2.445~$\pm$~0.204 & 20.56~$\pm$~0.315\\
    \bottomrule
    \end{tabular}}
    \label{tab:sota_compare}
    \end{subtable}
    \hfill%
    \begin{subtable}{0.49\textwidth}
    \centering
    \caption{SNN-RAT models vs clean models on various dataset.}
    \resizebox{\textwidth}{!}{
    \begin{tabular}{ll*{3}{c}}
    \toprule
     \multirow{2}{*}{Datasets} & \multirow{2}{*}{SNN variants} & \multicolumn{3}{c}{Validation accuracy per attack (\%)}\\
    \cmidrule{3-5}
     & & FGSM & PGD & Clean\\
    \midrule
    \multirow{5}{*}{CIFAR10} & SNN-RAT VGG-11 \cite{ding2022snn}$^{\dagger}$ & 45.23 & 21.16 & 90.74\\
    \cmidrule{2-5}
    & SNN-RAT WRN-16 \cite{ding2022snn}$^{\dagger}$ & 50.78 & 22.71 & \textbf{92.69}\\
    \cmidrule{2-5}
    & SEW-ResNet18 \cite{fang2021deep} & 46.99~$\pm$~0.285 & 34.33~$\pm$~0.386 & 73.78~$\pm$~0.351\\
    \cmidrule{2-5}
    & SEW-ResNet34 \cite{fang2021deep} & 49.20~$\pm$~0.421 & 36.98~$\pm$~0.312 & 74.16~$\pm$~0.451 \\
    \cmidrule{2-5}
    & SEW-ResNet50 \cite{fang2021deep} & \textbf{51.09~$\pm$~0.199} & \textbf{40.55~$\pm$~0.36} & 73.34~$\pm$~0.395 \\
    \midrule
    \multirow{5}{*}{CIFAR100} & SNN-RAT VGG-11 \cite{ding2022snn}$^{\dagger}$ & 25.86 & 10.38 & \textbf{70.89}\\
    \cmidrule{2-5}
    & SNN-RAT WRN-16 \cite{ding2022snn}$^{\dagger}$ & \textbf{28.08} & 11.31 & 69.32\\
    \cmidrule{2-5}
    & SEW-ResNet18 \cite{fang2021deep} & 20.81~$\pm$~0.523 & 13.47~$\pm$~0.246 & 42.19~$\pm$~0.373\\
    \cmidrule{2-5}
    & SEW-ResNet34 \cite{fang2021deep} & 19.91~$\pm$~0.427 & 13.25~$\pm$~0.382 & 41.81~$\pm$~0.292\\
    \cmidrule{2-5}
    & SEW-ResNet50 \cite{fang2021deep} & 21.27~$\pm$~0.272 & \textbf{16.03~$\pm$~0.292} & 38.91~$\pm$~0.387\\
    \midrule
    \multirow{5}{*}{CIFAR10-DVS} & SNN-RAT VGG-11 \cite{ding2022snn} & 36.22~$\pm$~0.374 & 19.02~$\pm$~0.296 & 71.77~$\pm$~0.272\\
    \cmidrule{2-5}
    & SNN-RAT WRN-16 \cite{ding2022snn} & 40.16~$\pm$~0.322 & 17.11~$\pm$~0.318 & 76.12~$\pm$~0.194 \\
    \cmidrule{2-5}
    & SEW-ResNet18 \cite{fang2021deep} & 47.19~$\pm$~0.467 & 21.47~$\pm$~0.416 & 75.18~$\pm$~0.398\\
    \cmidrule{2-5}
    & SEW-ResNet34 \cite{fang2021deep} & \textbf{49.03~$\pm$~0.288} & 18.65~$\pm$~0.183 & 75.29~$\pm$~0.304\\
    \cmidrule{2-5}
    & SEW-ResNet50 \cite{fang2021deep} & 48.52~$\pm$~0.441 & \textbf{22.34~$\pm$~0.277} & \textbf{76.53~$\pm$~0.373}\\
    \bottomrule
    \end{tabular}}
    \label{tab:dataset_compare}
    \end{subtable}
    \small$^{\dagger}$: Original accuracy from mentioned work.
    \label{fig:grad_compare}
    \vspace{-0.8\intextsep}
\end{table*}

The results shown in Table~\ref{tab:sota_compare} indicate that all SEW-ResNet variants achieve SOTA performance under PGD attacks, reaching up to 16.03\% accuracy, which represents a 4.72\% improvement over the previously best-performing method (SNN-RAT WRN-16). Under FGSM attacks, SEW-ResNet50 achieves 21.27\% accuracy, which is 4.59\% lower than SNN-RAT but remains comparable to Hire-SNN. This observation suggests that the reported benchmarking results may be influenced by obfuscated gradients \cite{athalye2018obfuscated}. In contrast, Spiking-ResNet models without explicit defense mechanisms do not demonstrate comparable robustness as deeper residual blocks are introduced. Consequently, subsequent benchmarking experiments focus solely on SEW-ResNet variants. As further illustrated in Table~\ref{tab:dataset_compare}, this robustness trend remains consistent across multiple datasets, with SEW-ResNet outperforming most existing adversarial defense methods: SEW-ResNet50 achieves a 0.31\% improvement under FGSM and a 17.84\% improvement under PGD compared with SNN-RAT WRN-16 on CIFAR10; SEW-ResNet50 improves PGD robustness by 4.72\% relative to SNN-RAT WRN-16 on CIFAR100; SEW-ResNet34 improves FGSM robustness by 8.87\% relative to SNN-RAT WRN-16 on CIFAR10-DVS, while SEW-ResNet50 improves PGD robustness by 3.32\% compared with SNN-RAT VGG-11 on CIFAR10-DVS.

\subsection{Gradient behavior analysis and theoretical interpretation}
\subsubsection{Analysis of gradient properties}

As shown in the benchmarking evaluation reported in Section~\ref{benchmark}, undefended SEW-ResNet models demonstrate a surprisingly strong degree of adversarial resilience, achieving performance comparable to several SOTA defense strategies. In contrast, Spiking-ResNet models do not exhibit the same level of robustness. This observation naturally leads to an important question:

\begin{itemize}
    \item \textit{What underlying mechanism enables SNNs to maintain robustness without explicitly designed defense techniques?}
\end{itemize}

We propose that this phenomenon is largely associated with the sparsity of input gradients in SEW-ResNet architectures. Sparse gradients at the input level tend to hinder adversarial optimization procedures, limiting the effectiveness of carefully structured perturbations while making the attack process more dependent on localized disturbances. Supporting this interpretation, previous studies indicate that the adversarial susceptibility of surrogate-trained SNNs is theoretically bounded by the density of input gradients \cite{liu2024enhancing}. More specifically, the ratio between adversarial vulnerability and random vulnerability grows proportionally with the number of non-zero entries in the gradient. Consequently, increased gradient sparsity inherently constrains the effectiveness of adversarial attacks.

\begin{figure*}[t!]
    \centering
    \begin{subfigure}{0.49\textwidth}
    \centerline{\includegraphics[width=\linewidth]{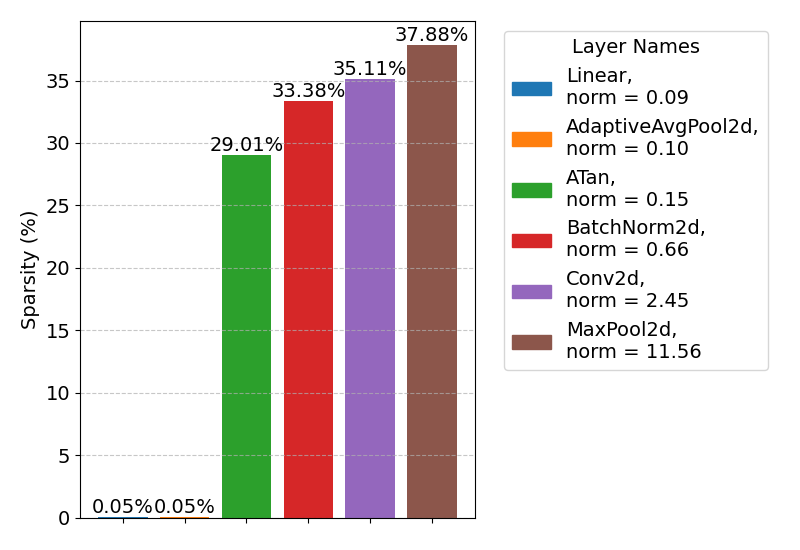}}
    \caption{Configuration with max pooling. Gradients remain well-defined while high sparsity mainly appears inside the SEW residual block operations.}
    \label{sparse_fig_sew}
    \end{subfigure}
    \hfill
    \begin{subfigure}{0.49\textwidth}
    \centerline{\includegraphics[width=\linewidth]{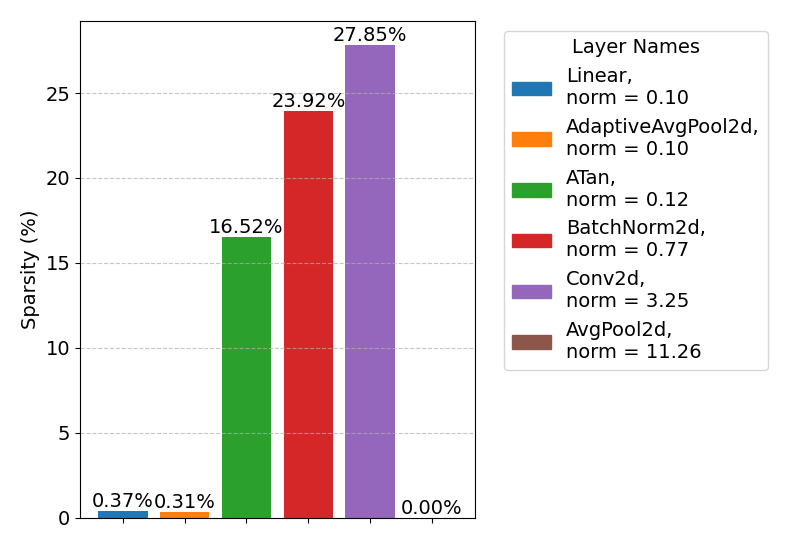}}
    \caption{Configuration with average pooling. Gradient sparsity is noticeably lower compared with the result in Figure~\ref{sparse_fig_sew}.}
    \label{sparse_fig_avg}
    \end{subfigure}
    \caption{Visualization of mean input gradient sparsity and gradient norm across the different operations within SEW-ResNet18.}
    \vspace{-\intextsep}
\end{figure*}

To examine this hypothesis, we conduct an empirical investigation of input gradient characteristics in SEW-ResNet models subjected to adversarial perturbations. We measure the average proportion of non-zero gradient elements (Fig.~\ref{sparse_fig_sew}) and additionally evaluate gradient norms to ensure that the observed sparsity does not arise from gradient masking or shattered gradient effects. This verification follows the evaluation guidelines proposed in \cite{athalye2018obfuscated}.

Although gradients remain properly defined, as confirmed by stable gradient norms, significant sparsity emerges primarily within residual block operations, ranging from 29.01\% in surrogate gradients of IF neurons to 37.88\% in gradients produced by max pooling layers. This effect is particularly prominent in convolution layers, batch normalization layers, max pooling operations, and the arctan surrogate gradient used in IF neurons. Such sparsity may function as a form of implicit regularization, weakening the effectiveness of gradient-based attacks and thereby contributing to the robustness observed in our experimental results.

\begin{equation}
    \begin{aligned}
        \frac{\partial \textbf{Conv}(x, \mathbf{w})}{\partial x} & = \textbf{Conv}^{\intercal}\left(\frac{\partial \mathcal{L}}{\partial y}, \mathbf{w}\right), 
        \quad \begin{aligned}
            \frac{\partial \textbf{BN}(y_{i,c})}{\partial x_{i,c}} & = \frac{\gamma_c}{\sqrt{\sigma_c^2 + \varepsilon}} \left[ \delta_{i,c} -  \sum_j^m \frac{\delta_{j,c}}{m} - \frac{\hat{x}_{i,c}}{m} \sum_j^m \delta_{j,c} \cdot \hat{x}_{j,c} \right],\\
        \delta_{i,c} &= \frac{\partial \mathcal{L}}{\partial y_{i,c}}, \quad
        \hat{x}_{i,c} = \frac{x_{i,c} - \mu_c}{\sqrt{\sigma_c^2 + \varepsilon}}, \quad y_{i,c} = \gamma_c \hat{x}_{i,c} + \beta_c\\
        \mu_c &= \frac{1}{m} \sum_{i=1}^{m} x_{i,c}, \quad \sigma_c^2 = \frac{1}{m} \sum_{i=1}^{m} (x_{i,c} - \mu_c)^2\\
        \end{aligned},\\
        \frac{\partial \textbf{MaxPool}(x)}{\partial x_{i,j}} & =
            \begin{cases}
                \frac{\partial \mathcal{L}}{\partial y_{p,q}} & \text{if } x_{i,j} = \max\limits_{(m,n) \in \mathcal{R}_{p,q}} x_{m,n} \\
                0 & \text{otherwise}
            \end{cases}, \quad \frac{\partial \Theta}{\partial x} = \frac{\alpha}{2\left( 1 + \left( \frac{\pi}{2} \alpha x \right)^2 \right)} \quad \text{where $\alpha=2$}\\
    \end{aligned}
    \label{eq:component}
\end{equation}

A deeper examination reveals that the observed gradient sparsity originates from two distinct mechanisms, which we categorize as:

\begin{itemize}
    \item \textit{Architectural sparsity}, where gradient sparsity arises from deliberate architectural design decisions.
    \item \textit{Natural sparsity}, where sparsity results from the intrinsic properties of spike-based signal processing rather than the network structure itself.
\end{itemize}

To better quantify these effects, we analyze the gradient formulations of each operation within the spike-based element-wise residual blocks. As indicated in Equation~\ref{eq:component}, the gradients associated with convolution \textbf{Conv}, batch normalization \textbf{BN}, and the surrogate gradient term $\Theta$ depend continuously on both input and intermediate activations. In contrast, the max pooling operation \textbf{MaxPool} introduces a structural constraint: gradients remain non-zero only at positions corresponding to maximum activations within each receptive field, while all other positions receive zero gradients.

These zero-gradient regions propagate backward during training, generating sparsity in earlier layers. We refer to this phenomenon as \textit{architectural sparsity}. In SEW-ResNet architectures, this effect is further amplified due to the strongly coupled nature of residual blocks. Consequently, substituting max pooling with an alternative operation that preserves gradient flow may increase gradient density.

To evaluate this hypothesis, we replace max pooling \textbf{MaxPool} with average pooling \textbf{AvgPool}, whose gradient is expressed as 
$\frac{\partial \textbf{AvgPool}}{\partial x_{i,j}} = \sum_{(p,q) \;:\; (i,j) \in \mathcal{R}_{p,q}} \;
\frac{1}{|\mathcal{R}_{p,q}|} \cdot
\frac{\partial \mathcal{L}}{\partial y_{p,q}}$. 
This formulation evenly distributes gradient contributions across all elements within the receptive region \( \mathcal{R}_{p,q} \), thereby reducing gradient sparsity and increasing the influence of each element during gradient-based optimization.

We subsequently reevaluate the SEW-ResNet variants under this modified setting and measure both the resulting input gradient sparsity and adversarial robustness. The corresponding results are presented in Figure~\ref{sparse_fig_avg} and Table~\ref{tab:rebenchmark}.

From Table~\ref{tab:rebenchmark}, it is evident that adversarial robustness decreases when average pooling replaces max pooling. Under FGSM and PGD attacks, validation accuracy drops significantly, reaching as low as 0.16\% and 2.88\% for SEW-ResNet34 on CIFAR-100, and at best 1.28\% and 8.61\% for SEW-ResNet18 on CIFAR-10. Conversely, performance on clean validation data improves, with increases ranging from 1.13\% (SEW-ResNet18) to 4.49\% (SEW-ResNet50) on CIFAR-100.
\begin{table}[t!]
    \centering
    \caption{Validation accuracy comparison on CIFAR-10 and CIFAR-100 for undefended SEW-ResNet variants \cite{fang2021deep} when average pooling is employed. The highest accuracy for each attack scenario on a given dataset is highlighted in bold. The adversarial attack configurations follow the same settings used in Table~\ref{tab:sota_compare}. Accuracy decreases relative to Table~\ref{tab:dataset_compare} are indicated in red, while improvements are shown in green. The results reveal a clear trade-off between clean accuracy and adversarial robustness.}
    \resizebox{\linewidth}{!}{
    \begin{tabular}{l*{11}{c}}
    \toprule
    \multirow{2}{*}{Attack} & \multicolumn{3}{c}{CIFAR-10} & & \multicolumn{3}{c}{CIFAR-100} & & \multicolumn{3}{c}{CIFAR10-DVS} \\
    \cmidrule{2-4}\cmidrule{6-8}\cmidrule{10-12}
    & SEW-ResNet18 & SEW-ResNet34 & SEW-ResNet50 & & SEW-ResNet18 & SEW-ResNet34 & SEW-ResNet50 & & SEW-ResNet18 & SEW-ResNet34 & SEW-ResNet50 \\
    \midrule
    FGSM & 45.71~$\pm$~0.257 \textcolor{red}{($\downarrow$1.28)} & 48.72~$\pm$~0.581 \textcolor{red}{($\downarrow$0.48)} & \textbf{49.88~$\pm$~0.374} \textcolor{red}{($\downarrow$1.21)} & & 17.93~$\pm$~0.364 \textcolor{red}{($\downarrow$2.88)} & 20.07~$\pm$~0.425 \textcolor{red}{($\downarrow$0.16)} & \textbf{22.02~$\pm$~0.476} \textcolor{red}{($\downarrow$0.74)} & & 46.27~$\pm$~0.377 \textcolor{red}{($\downarrow$0.92)} & \textbf{47.66~$\pm$~0.317} \textcolor{red}{($\downarrow$1.37)} & 47.01~$\pm$~0.258 \textcolor{red}{($\downarrow$1.51)} \\
    \midrule
    PGD & 25.72~$\pm$~0.395 \textcolor{red}{($\downarrow$8.61)} & 29.82~$\pm$~0.624 \textcolor{red}{($\downarrow$7.16)} & \textbf{33.29~$\pm$~0.273} \textcolor{red}{($\downarrow$7.26)} & & 8.65~$\pm$~0.219 \textcolor{red}{($\downarrow$4.82)} & 10.37~$\pm$~0.388 \textcolor{red}{($\downarrow$2.88)} & \textbf{13.37~$\pm$~0.451} \textcolor{red}{($\downarrow$2.66)} & & 13.94~$\pm$~0.276 \textcolor{red}{($\downarrow$7.53)} & 12.54~$\pm$~0.402 \textcolor{red}{($\downarrow$6.11)} & \textbf{14.30~$\pm$~0.442} \textcolor{red}{($\downarrow$8.04)} \\
    \midrule
    Clean & 75.76~$\pm$~0.286 \textcolor{green}{($\uparrow$1.98)} & \textbf{76.17~$\pm$~0.398} \textcolor{green}{($\uparrow$2.01)} & 75.28~$\pm$~0.397 \textcolor{green}{($\uparrow$1.94)} & & 43.32~$\pm$~0.377 \textcolor{green}{($\uparrow$1.13)} & \textbf{44.58~$\pm$~0.634} \textcolor{green}{($\uparrow$2.77)} & 43.40~$\pm$~0.321 \textcolor{green}{($\uparrow$4.49)} & & 77.77~$\pm$~0.308 \textcolor{green}{($\uparrow$2.59)} & 76.43~$\pm$~0.358 \textcolor{green}{($\uparrow$1.14)} & \textbf{78.35~$\pm$~0.508} \textcolor{green}{($\uparrow$1.82)} \\
    \bottomrule
    \end{tabular}}
    \label{tab:rebenchmark}
    \vspace{-0.9\intextsep}
\end{table}
Figure~\ref{sparse_fig_avg} illustrates that substituting max pooling with average pooling leads to a substantial decrease in input gradient sparsity across all layers. In particular, sparsity introduced by the pooling layer is completely eliminated, corresponding to a 37.88\% reduction relative to the max pooling configuration. Meanwhile, spiking neuron layers still exhibit sparsity levels as low as 16.52\%. These findings are consistent with previous studies highlighting the importance of gradient magnitude in optimizing SNNs \cite{fang2021deep} and the effectiveness of gradient scaling in improving model generalization \cite{11261679}. However, entirely eliminating gradient sparsity is theoretically infeasible due to the inherently sparse firing behavior of spiking neurons, commonly modeled by the Heaviside activation function (Eq.~\ref{heviside}). This phenomenon is referred to as \textit{natural sparsity}.

\subsubsection{Theoretical explanation}

Prior research has established a strong connection between the magnitude of parameter gradients and the performance of neural models \cite{11261679, fang2021deep, shalev2017failures}. In particular, \cite{11261679} demonstrated that the performance of surrogate-trained SNNs is closely related to the scaling of gradients across the temporal dimension $T$, characterized through the gradient signal-to-noise ratio (SNR). Motivated by these findings, we investigate how sparsity in the input gradients may attenuate parameter gradients during optimization and consequently influence model performance. To formalize this intuition, we introduce the following theorem:

\begin{theorem}
Given a multilayer SNN of the form:
\begin{equation}
    p_\mathbf{w}(s) = \mathbb{E}_T\left[y_{\mathbf{w}_{2}}(y_{\mathbf{w}_{1}}(s[t]))\right] \quad \text{where} \quad t \in T, \quad
    \begin{aligned}
    & y_{\mathbf{w}_{1}}(\mathbf{x}) = \Theta(z_{\mathbf{w}_{1}}(\mathbf{x})), \quad z_{\mathbf{w}_{1}}(\mathbf{x}) = \mathbf{w}_{1} \cdot \mathbf{x}\\
    & y_{\mathbf{w}_{2}}(\mathbf{x}) = \Theta(z_{\mathbf{w}_{2}}(\mathbf{x})), \quad z_{\mathbf{w}_{2}}(\mathbf{x}) = \mathbf{w}_2 \cdot \mathbf{x}\\
    \end{aligned}
    \label{eq:example}
\end{equation}
where $s[t] \in \{0, 1\}^{T \times d}$ denotes the input spike signal, $\mathbf{w}_l \in \mathbb{R}^{d \times d}$ represents the parameter matrix at layer $l$, and $\Theta(x, V_{th})$ corresponds to the hard-reset IF neuron model with membrane threshold $V_{th}$, defined similarly to Equation~\ref{heviside}. Under this formulation, the following observations hold:
\begin{itemize}
    \item Any sparsity appearing within the shared gradient propagation path will propagate backward to both the input gradient and the weight gradients.
    \item Parameters located closer to the input signal share a longer gradient chain, implying that sparsity in the input gradient has a stronger influence on their parameter gradients.
\end{itemize}
\label{theorem:eq}
\end{theorem}

\begin{proof}
The gradient of the model output with respect to the input signal $s[t]$ at timestep $t$, denoted as $\nabla_{s[t]} p_\mathbf{w}(s)$, can be expressed as:
\[
    \begin{aligned}
        \nabla_{s[t]} p_\mathbf{w}(s) & = \frac{\partial p_\mathbf{w}}{\partial y_{\mathbf{w}_{2}}} \cdot \frac{\partial y_{\mathbf{w}_{2}}}{\partial z_{\mathbf{w}_{2}}}\cdot \frac{\partial z_{\mathbf{w}_{2}}}{\partial y_{\mathbf{w}_{1}}} \cdot \frac{\partial y_{\mathbf{w}_{2}}}{\partial z_{\mathbf{w}_{1}}}\cdot \frac{\partial z_{\mathbf{w}_{1}}}{\partial s[t]}  = \frac{1}{T} \cdot \frac{\mathbf{w}_{2}}{ 1 + \left( \pi \cdot \mathbf{w}_{2}\cdot \Theta(\mathbf{w}_{1} \cdot s[t]) \right)^2} \cdot \frac{ \mathbf{w}_{1}}{ 1 + \left( \pi \cdot \mathbf{w}_{1}\cdot s[t] \right)^2}.\\
    \end{aligned}
\]
Similarly, the cumulative Jacobian of the model with respect to the parameters $\mathbf{w}_l$ at layer $l$, denoted $\nabla_{\mathbf{w}_l} p_\mathbf{w}(s)$, can be written as:
\begin{equation}
    \begin{aligned}
    & \nabla_{\mathbf{w}_2} p_\mathbf{w}(s) = \frac{\partial p_\mathbf{w}}{\partial y_{\mathbf{w}_2}} \cdot \frac{\partial y_{\mathbf{w}_2}}{\partial z_{\mathbf{w}_2}}\cdot \frac{\partial z_{\mathbf{w}_2}}{\partial \mathbf{w}_2} = \mathbb{E}_T\left[\frac{\Theta(\mathbf{w}_{1} \cdot s[t])}{ 1 + \left( \pi \cdot\mathbf{w}_{2} \cdot \Theta(\mathbf{w}_{1} \cdot s[t]) \right)^2}\right]\\
    & \begin{aligned}
        \nabla_{\mathbf{w}_1} p_\mathbf{w}(s) & = \frac{\partial p_\mathbf{w}}{\partial y_{\mathbf{w}_2}} \cdot \frac{\partial y_{\mathbf{w}_2}}{\partial z_{\mathbf{w}_2}}\cdot \frac{\partial z_{\mathbf{w}_2}}{\partial y_{\mathbf{w}_1}} \cdot \frac{\partial y_{\mathbf{w}_1}}{\partial z_{\mathbf{w}_1}}\cdot \frac{\partial z_{\mathbf{w}_1}}{\partial \mathbf{w}_1} = \mathbb{E}_T\left[ \frac{\mathbf{w}_{2}}{ 1 + \left( \pi \cdot \mathbf{w}_{2}\cdot \Theta(\mathbf{w}_{1} \cdot s[t]) \right)^2} \cdot \frac{s[t]}{ 1 + \left( \pi \cdot \mathbf{w}_{1}\cdot s[t] \right)^2}\right]\\
    \end{aligned}\\
    \end{aligned}
\end{equation}
From these derivations, we observe that the input gradient $\nabla_{s[t]} p_\mathbf{w}(s)$ shares a substantial portion of the gradient propagation chain with the Jacobian $\nabla_{\mathbf{w}_1} p_\mathbf{w}(s)$ associated with $\mathbf{w}_1$. Specifically, both quantities include the sequence of terms 
$\frac{\partial p_\mathbf{w}}{\partial y_{\mathbf{w}_2}} \cdot \frac{\partial y_{\mathbf{w}_2}}{\partial z_{\mathbf{w}_2}}\cdot \frac{\partial z_{\mathbf{w}_2}}{\partial y_{\mathbf{w}_1}} \cdot \frac{\partial y_{\mathbf{w}_1}}{\partial z_{\mathbf{w}_1}}$
during backpropagation. In contrast, the gradient with respect to $\mathbf{w}_2$ depends on a shorter chain consisting only of 
$\frac{\partial p_\mathbf{w}}{\partial y_{\mathbf{w}_2}} \cdot \frac{\partial y_{\mathbf{w}_2}}{\partial z_{\mathbf{w}_2}}$.
These relationships motivate the observations discussed earlier.\footnote{A graphical illustration of these gradient propagation paths is provided in Figure \ref{fig:grad_path}.}

\begin{figure}[t!]
\centerline{\includegraphics[width=\linewidth]{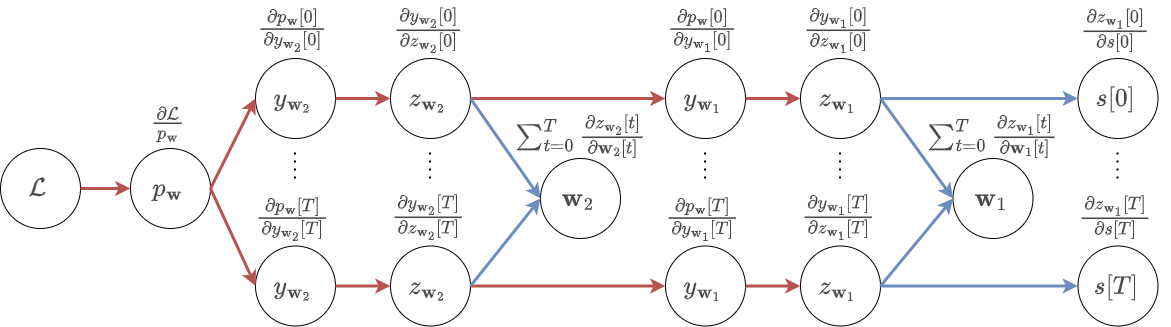}}
\caption{Backpropagation structure corresponding to Equation \ref{eq:example}. Red arrows represent gradient components shared across different derivative paths, whereas blue arrows indicate gradients that are specific to individual variables. Sparsity arising along shared gradient paths propagates to both the input and trainable parameters.}
\label{fig:grad_path}
\end{figure}
\end{proof}

From this result, it follows that sparsity occurring in the shared gradient path inevitably affects both input and parameter gradients during backpropagation. However, beyond the qualitative effect of sparsity, an important question remains: how can we quantitatively characterize the relationship between the density of weight gradients (particularly those of $\mathbf{w}_1$, which is closest to the input and most strongly influenced by the shared gradient path) and the density of input gradients under the conditions described in Theorem \ref{theorem:eq}? To address this, we present the following theorem, which formally describes this relationship.

\begin{theorem}
Under the same conditions as Theorem \ref{theorem:eq}, consider a parameter matrix $\mathbf{w}_l \in \mathbb{R}^{d_1 \times d_2}$ at layer $l$. The matrix is said to possess a row/column density degree of $k^{(i)}_{\mathbf{w}_l}$/$k^{(j)}_{\mathbf{w}_l}$ if each row or column vector $\mathbf{w}_l^{(i)}/\mathbf{w}_l^{(j)}$ (where $i \in d_1$, $j \in d_2$) contains at most $k$ non-zero elements $\mathbf{w}^{(ij)}_l$. Formally, this is defined as:
\begin{equation}
    \begin{aligned}
    k^{(i)}_{\mathbf{w}_l} & = \sup_{i \in d_1}\left(\#\{ \mathbf{w}^{(ij)}_l \in \mathbf{w}^{(i)}_l | \mathbf{w}^{(ij)}_l \neq 0\}\right) = \sup_{i \in d_1}\left(\#\text{supp}\left(\mathbf{w}^{(i)}_l\right)\right) = \sup_{i \in d_1}\left(\|\mathbf{w}^{(i)}_l\|_0\right),\\
    k^{(j)}_{\mathbf{w}_l} & = \sup_{j \in d_2}\left(\#\{ \mathbf{w}^{(ij)}_l \in \mathbf{w}^{(j)} | \mathbf{w}^{(ij)}_l \neq 0\}\right) = \sup_{j \in d_2}\left(\#\text{supp}\left(\mathbf{w}^{(j)}_l\right)\right) = \sup_{j \in d_2}\left(\|\mathbf{w}^{(j)}_l\|_0\right),\\
    \end{aligned}
\end{equation}

and for a vector $\mathbf{v} \in \mathbb{R}^{d}$ we define
$k_{\mathbf{v}} = \sup_{i \in d}\left(\#\{ \mathbf{v}^{(i)} \in \mathbf{v} | \mathbf{v}^{(i)} \neq 0\}\right) = \sup_{i \in d}\left(\#\text{supp}\left(\mathbf{v}\right)\right) = \sup_{i \in d}\left(\|\mathbf{v}\|_0\right)$.

If the $L_0$ norms of the input gradient and weight gradient are defined as:
\begin{equation}
     \| \nabla_{s[t]} p_\mathbf{w}(s) \|_0 = \| \mathbf{w}_{2} \cdot \mathbf{w}_{1} \|_0,\quad \| \nabla_{\mathbf{w}_1} p_\mathbf{w}(s) \|_0 =  \sum_{t=1}^{T}{\|\mathbf{w}_{2} \cdot s[t]} \|_0= \| \mathbf{w}_{2} \cdot S \|_0 \quad (S \in \{0,1\}^{d\times T}),
\end{equation}

and all weights satisfy $k^{(i)}_{\mathbf{w}_l}>0, k^{(j)}_{\mathbf{w}_l}>0$ after optimization, then the $L_0$ norm of the weight gradient for $\mathbf{w}_1$ is bounded by

\begin{equation}
        \mathbb{E}\left[ \| \nabla_{\mathbf{w}_1} p_\mathbf{w}(s) \|_0 \right] \leq \frac{T}{d} \cdot \frac{k^{(j)}_{S}}{k^{(j)}_{\mathbf{w}_1}} \cdot \mathbb{E}\left[\| \nabla_{s[t]} p_\mathbf{w}(s)\|_0 \right].
\end{equation}

\label{theorem:main}
\end{theorem}

\begin{proof}
Each element of the input gradient can be expressed as $\nabla_{s[t]} p_\mathbf{w}(s)^{(ij)}  = \sum_{n=1}^{d} \mathbf{w}^{(in)}_{2} \cdot \mathbf{w}^{(nj)}_{1}$. The probability that an individual term in this summation chain, $\mathbf{w}^{(in)}_{2} \cdot \mathbf{w}^{(nj)}_{1}$, is non-zero can be upper bounded by $Pr\left(\mathbf{w}^{(in)}_{2} \cdot \mathbf{w}^{(nj)}_{1} \neq 0\right) \leq \frac{k^{(i)}_{\mathbf{w}_2}k^{(j)}_{\mathbf{w}_1}}{d^2}$. Since the gradient entry becomes non-zero whenever at least one element in the summation chain is non-zero, the probability that a particular entry of the input gradient matrix $\nabla_{s[t]} p_\mathbf{w}(s)^{(ij)}$ is non-zero can be bounded as $Pr\left(\nabla_{s[t]} p_\mathbf{w}(s)^{(ij)} \neq 0 \right) \leq\frac{k^{(i)}_{\mathbf{w}_2}k^{(j)}_{\mathbf{w}_1}}{d^2}$. Consequently, the expected $L_0$ norm of the input gradient matrix over a $d \times d$ dimensional space can be bounded by
\begin{equation}
    \mathbb{E}\left[\| \nabla_{s[t]} p_\mathbf{w}(s)\|_0 \right] \leq k^{(i)}_{\mathbf{w}_2}k^{(j)}_{\mathbf{w}_1}.
    \label{eq:res1}
\end{equation}
An analogous derivation can be performed for the gradient with respect to the parameter $\mathbf{w}_1$, yielding an expectation bound for its $L_0$ norm. Considering a single term in the summation chain $\sum_{n=1}^{d} \mathbf{w}^{(in)}_{2} \cdot S^{(nj)}$, we obtain:
\begin{equation}
    Pr\left( \mathbf{w}^{(in)}_{2} \cdot S^{(nj)} \neq 0\right) \leq \frac{k^{(i)}_{\mathbf{w}_2}k^{(j)}_{S}}{d^2}, \quad \Leftrightarrow \quad  \mathbb{E}\left[\| \nabla_{\mathbf{w}_1} p_\mathbf{w}(s) \|_0 \right] \leq T \cdot \frac{k^{(i)}_{\mathbf{w}_2}k^{(j)}_{S}}{d}.
    \label{eq:res2}
\end{equation}
Dividing the inequalities obtained from Equation \ref{eq:res1} and Equation \ref{eq:res2} gives:
\begin{equation}
    \begin{aligned}
        & \frac{\mathbb{E}\left[ \| \nabla_{\mathbf{w}_1} p_\mathbf{w}(s) \|_0 \right]}{\mathbb{E}\left[\| \nabla_{s[t]} p_\mathbf{w}(s)\|_0 \right]} \leq \frac{T \cdot \frac{k^{(i)}_{\mathbf{w}_2}k^{(j)}_{S}}{d}}{k^{(i)}_{\mathbf{w}_2}k^{(j)}_{\mathbf{w}_1}},\quad
        \Leftrightarrow \quad  \mathbb{E}\left[ \| \nabla_{\mathbf{w}_1} p_\mathbf{w}(s) \|_0 \right] \leq \frac{T}{d} \cdot \frac{k^{(i)}_{\mathbf{w}_2}k^{(j)}_{S}}{k^{(i)}_{\mathbf{w}_2}k^{(j)}_{\mathbf{w}_1}} \cdot \mathbb{E}\left[\| \nabla_{s[t]} p_\mathbf{w}(s)\|_0 \right],\\
        & \Leftrightarrow \quad \mathbb{E}\left[ \| \nabla_{\mathbf{w}_1} p_\mathbf{w}(s) \|_0 \right] \leq \frac{T}{d} \cdot \frac{k^{(j)}_{S}}{k^{(j)}_{\mathbf{w}_1}} \cdot \mathbb{E}\left[\| \nabla_{s[t]} p_\mathbf{w}(s)\|_0 \right].\\
    \end{aligned}
\end{equation}
These results indicate that the density of the weight gradient is bounded by the density of the input gradient, scaled by a factor of $ \frac{T}{d} \cdot \frac{k^{(j)}_{S}}{k^{(j)}_{\mathbf{w}_1}}$.
\end{proof}

When combined with the observations from \cite{liu2024enhancing}, these results highlight a clear chain of dependencies connecting weight gradients, input gradients, and adversarial robustness. In particular, strategies that regularize the $L_0$ norm of input gradients to improve robustness (as explored in \cite{liu2024enhancing}) may also reduce the magnitude of weight gradients, thereby influencing the optimization dynamics.

\section{Limitations and future works}

Although our study provides strong empirical and theoretical evidence linking gradient density–sparsity to the trade-off between adversarial robustness and generalization in SNNs, it does not introduce a unified method that simultaneously improves both properties. Nevertheless, our findings open several promising research directions. One potential avenue is sparsity-aware architectural design, where future SNN architectures explicitly balance natural sparsity and architecture-induced sparsity to achieve a more favorable trade-off between robustness and accuracy. This may involve hybrid pooling strategies or novel spike-based normalization layers designed to preserve beneficial sparsity patterns. Another possible direction is the development of stronger attack methods specifically tailored for SNNs, particularly for models with extremely sparse gradients where conventional white-box gradient-based attacks become ineffective.

\section{Conclusion}

In this work, we systematically investigated the trade-off between adversarial robustness and generalization in undefended spiking neural networks. Through extensive benchmarking across multiple datasets, we observed that these models exhibit strong robustness against conventional white-box attacks, often outperforming existing state-of-the-art defense mechanisms despite the absence of explicit adversarial training or defense strategies. To understand the origin of this behavior, we performed an empirical analysis of input gradient distributions in SNNs and identified significant gradient sparsity across multiple network operations. We further categorized these sparsity patterns into two forms: \textit{architectural sparsity}, introduced by structural components of the network, and \textit{natural sparsity}, which arises from the intrinsic spike-based dynamics of SNNs. Experimental modifications that reduced gradient sparsity resulted in a measurable decline in adversarial robustness while improving clean accuracy. Complementary theoretical analysis further supports the existence of this trade-off, revealing a fundamental relationship between gradient sparsity, robustness, and generalization in spiking neural networks.

\section*{Data Availability Statement}
Source code for this research can be found at \url{https://github.com/luutn2002/grad-obf-snn}.

\section*{Author Contribution Statement}
All author contributed equally to the paper.

\bibliographystyle{vancouver}
\bibliography{refs}

\end{document}